\documentclass[10pt,twocolumn,letterpaper]{article}

\usepackage[pagenumbers]{iccv} % To force page numbers, e.g. for an arXiv version

\definecolor{iccvblue}{rgb}{0.21,0.49,0.74}
\usepackage[pagebackref,breaklinks,colorlinks,allcolors=iccvblue]{hyperref}

\usepackage[utf8]{inputenc} % allow utf-8 input
\usepackage[T1]{fontenc}    % use 8-bit T1 fonts
\usepackage{hyperref}       % hyperlinks
\usepackage{url}            % simple URL typesetting
\usepackage{booktabs}       % professional-quality tables
\usepackage{amsfonts}       % blackboard math symbols
\usepackage{nicefrac}       % compact symbols for 1/2, etc.
\usepackage{microtype}      % microtypography
\usepackage{xcolor}         % colors
\usepackage{dsfont}
\usepackage{graphicx}
\usepackage{amsmath}
\usepackage{color}
\usepackage{colortbl}
\usepackage{bbding}
\usepackage{graphicx, amsmath, amssymb, caption, subcaption, multirow, overpic, textpos}
\usepackage{tabulary}
\definecolor{Gray}{gray}{0.95}
\newcommand{\gr}[1]{{\textcolor{gray}{#1}}}
\newlength\savewidth
\definecolor{baselinecolor}{gray}{.95}

\def\eg{\emph{e.g.}}

\def\paperID{2113} % *** Enter the Paper ID here
\def\confName{ICCV}
\def\confYear{2025}

\title{Unified Open-World Segmentation with Multi-Modal Prompts}

\author{
Yang Liu$^{1}$\thanks{Equal contribution. $^{\dag}$HC is the corresponding author.} \quad
Yufei Yin$^{2}$\textsuperscript{*} \quad
Chenchen Jing$^{3}$ \quad
Muzhi Zhu$^{1}$ \quad
Hao Chen$^{1\dag}$ \quad
Yuling Xi$^{1}$ \\[0.3em]
Bo Feng$^{4}$ \quad
Hao Wang$^{4}$ \quad
Shiyu Li$^{4}$ \quad
Chunhua Shen$^{1}$ \\[0.25cm]
{\normalsize
$^{1}$ Zhejiang University \quad
$^{2}$ Hangzhou Dianzi University \quad
$^{3}$ Zhejiang University of Technology \quad
$^{4}$ Apple
}
}

\begin{document}

\makeatletter
\let\@oldmaketitle\@maketitle%
\renewcommand{\@maketitle}{\@oldmaketitle%
\vspace{-2em}
    \centering
    \includegraphics[width=1\linewidth]{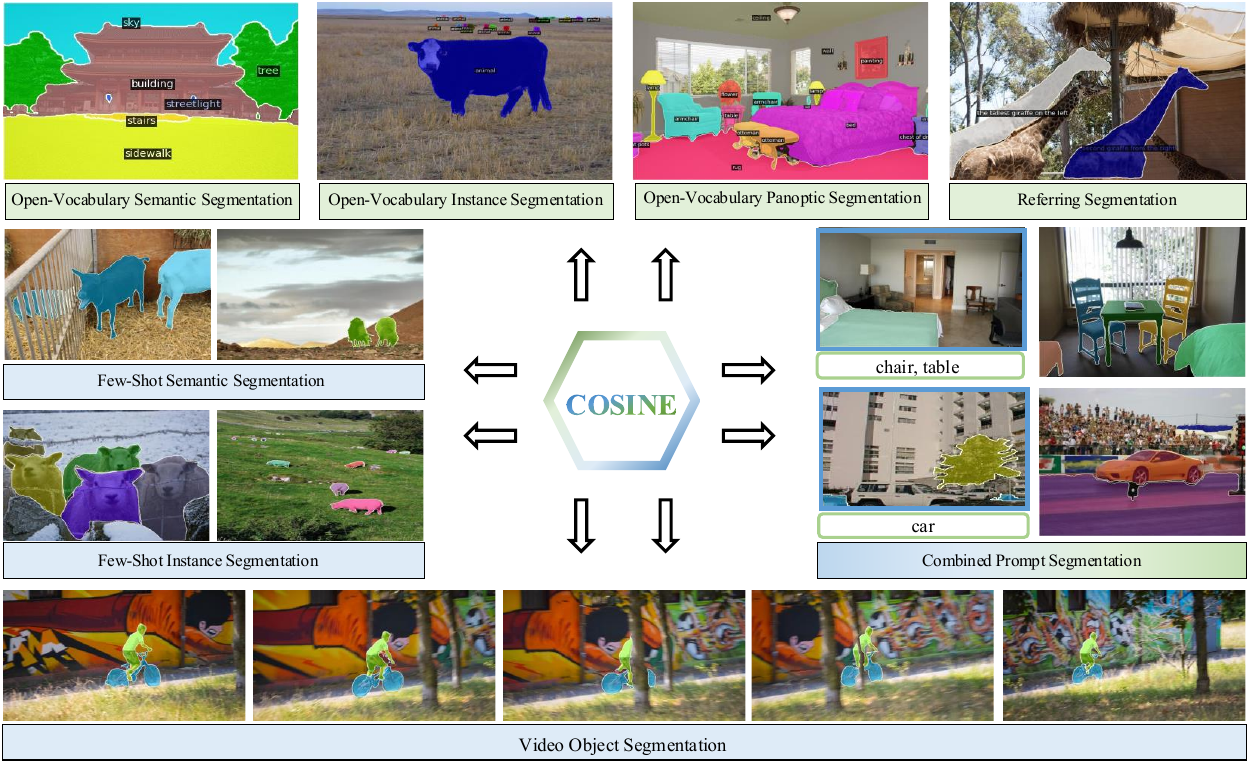}
    \captionof{figure}{COSINE is a unified open-world segmentation model that consolidates open-vocabulary and in-context segmentation. COSINE can simultaneously support text prompts (\textcolor{green}{green} boxes) and image prompts (\textcolor{blue}{blue} boxes) as inputs to perform various segmentation tasks, including semantic segmentation, instance segmentation, panoptic segmentation, referring segmentation, and video object segmentation. In addition, COSINE can collaboratively use different types of prompts to perform various segmentation tasks.
    }
    \label{fig:teaser}
   \bigskip}                   %
\makeatother

\maketitle

\begin{abstract}

In this work, we present COSINE, a unified open-world segmentation model that \textbf{C}onsolidates \textbf{O}pen-vocabulary \textbf{S}egmentation and \textbf{IN}-context s\textbf{E}gmentation with multi-modal prompts (e.g. text and image). 
COSINE exploits foundation models to extract representations for an input image and corresponding multi-modal prompts, and a SegDecoder to align these representations, model their interaction, and obtain masks specified by input prompts across different granularities.
In this way, COSINE overcomes architectural discrepancies, divergent learning objectives, and distinct representation learning strategies of previous pipelines for open-vocabulary segmentation and in-context segmentation.
Comprehensive experiments demonstrate that COSINE has significant performance improvements in both open-vocabulary and in-context segmentation tasks. 
Our exploratory analyses highlight that the synergistic collaboration between using visual and textual prompts leads to significantly improved generalization over single-modality approaches. Our code is released at \url{https://github.com/aim-uofa/COSINE}.

\end{abstract}

\section{Introduction}
\label{sec:intro}

Image segmentation~\cite{zhou2019semantic,lin2014microsoft,kirillov2019panoptic} aims to provide accurate conceptual localization, enabling precise understanding and content analysis of images at the pixel level.
Recently, image segmentation has gradually evolved 
from closed-world scenarios \cite{long2015fully,de2017semantic,he2017mask,zhao2017pyramid,wang2021solo,cheng2022masked} to open-world settings \cite{kirillov2023segment,zhu2023segprompt,xu2023open,yu2023convolutions,wang2023seggpt,liu2023matcher,li2024visual}. 
Unlike traditional closed-world segmentation models, which are limited to recognizing a fixed set of categories encountered during training, open-world segmentation models can localize arbitrary relevant objects in the wild based on user-provided prompts. 
Such models can enhance adaptability and robustness in unpredictable dynamic environments, such as autonomous driving and interactive robotics.

The current landscape of open-world segmentation research primarily revolves around two distinct paradigms: (1) Open-vocabulary segmentation~\cite{ghiasi2022scaling,liang2023open,xu2023side,xu2023open,yu2023convolutions,zhang2023simple}, which replaces learnable classifiers with textual embeddings derived from category descriptors, thereby extending conventional closed-set segmentation frameworks to recognize novel categories through natural language alignment; and (2) In-context segmentation~\cite{wang2023seggpt,liu2023matcher,zhang2023personalize,li2024visual,liu2024simple,meng2024segic}, which leverages contextual cues from example images to facilitate adaptive object segmentation in query images.
Despite significant progress in both directions, current methods predominantly address these tasks in isolation, failing to holistically tackle the complexities of open-world segmentation. 
Relying solely on text may lead to insufficient fine-grained semantic abstraction, whereas image-based exemplars often lack explicit category boundaries and semantic alignment. This raises a critical question: \textit{Can we unify open-vocabulary segmentation and in-context segmentation within a single framework, leveraging the complementary strengths of textual and visual modalities to enhance open-world segmentation capabilities?} 

We compare these two paradigms and identify the following problems in unifying them:
1) Architectural Discrepancies: Existing methods exhibit substantial structural differences. For instance, SegGPT~\cite{wang2023seggpt} employs a ViT-like~\cite{dosovitskiy2020image} encoder-only architecture for in-context segmentation, whereas ODISE~\cite{xu2023open} adopts a Mask2Former-like~\cite{cheng2022masked} encoder-decoder structure for open-vocabulary segmentation. 
2) Divergent Learning Objectives: Open-vocabulary segmentation primarily focuses on image-text semantic alignment, aiming to learn the association between images and class descriptions for recognizing novel categories in an open-world setting. In contrast, in-context segmentation emphasizes reference-query relationship modeling, leveraging contextual cues from example images to achieve adaptive target segmentation.
3) Distinct Representation Learning Strategies: Open-vocabulary segmentation typically relies on multimodal models (e.g., CLIP~\cite{radford2021learning}, Stable Diffusion~\cite{rombach2022high}) to leverage text embeddings for category matching, whereas in-context segmentation predominantly utilizes vision foundation models (e.g., MAE~\cite{he2022masked}, DINOv2~\cite{oquab2023dinov2}), performing target localization on visual features.
These fundamental differences pose challenges in designing a unified framework that effectively integrates both paradigms while preserving their advantages.

To address these challenges, we present COSINE, a unified open-world segmentation model that \textbf{C}onsolidates \textbf{O}pen-vocabulary \textbf{S}egmentation and \textbf{IN}-context s\textbf{E}gmentation.
Specifically, we first deploy a frozen Model Pool consisting of multiple foundation models to extract representations for the target image and different modality prompts (e.g., text and image), and convert them into token sequences. This standardization of the input format facilitates structural unification across both tasks, thereby enabling a single decoder-only segmentation model, named SegDecoder, to jointly process them.
The SegDecoder contains an Image-Prompt Aligner module and a Multi-Modality Decoder.
The Image-Prompt Aligner module aligns the image with various prompts, reducing the modality gap and learning a unified multi-modal representation space.
The Multi-Modality Decoder is used to model the interaction between object queries, the image, and different modality prompts, enabling the effective generation of masks at different granularities (e.g., semantic, instance).
COSINE optimizes the lightweight SegDecoder only, effectively unleashing the potential of the foundation models for open-world segmentation. 
As illustrated in Fig.~\ref{fig:teaser}, COSINE can concurrently address open-vocabulary and in-context segmentation tasks at multiple granularities, including semantic, instance, and panoptic segmentation.
Comprehensive experiments demonstrate that COSINE effectively unifies both settings within a single model, achieving state-of-the-art performance. In particular, our exploratory analyses highlight the synergistic collaboration between the visual and textual branches, leading to significantly improved generalization over single-modality approaches. We believe our findings offer valuable insights to the research community.

Our main contributions are as follows:
(1) To our knowledge, our method is the first to unify in-context segmentation and open-vocabulary segmentation. We present a simple but effective framework, COSINE, which unleashes the potential of frozen foundation models across various segmentation tasks.
(2) Our comprehensive experiments demonstrate that COSINE achieves significant performance improvements in both open-vocabulary and in-context segmentation tasks. 
(3) We further observe that the synergistic collaboration between different modality branches enhances generalization in open-world segmentation, providing valuable insights for the research community.

\section{Related Work}
\label{sec:related}

\noindent\textbf{Open-World Segmentation.}
Image segmentation~\cite{zhou2019semantic, lin2014microsoft, kirillov2019panoptic}, which aims to localize and organize meaningful concepts at the pixel level, has evolved from closed-world scenarios~\cite{long2015fully, de2017semantic, he2017mask, zhao2017pyramid, wang2021solo, cheng2022masked} to open-world settings~\cite{kirillov2023segment, zhu2023segprompt, xu2023open, yu2023convolutions, wang2023seggpt, liu2023matcher, li2024visual}. Open-world segmentation can be broadly categorized into two paradigms:
Open-vocabulary segmentation~\cite{ghiasi2022scaling,liang2023open,xu2023side,xu2023open,yu2023convolutions,zhang2023simple} focuses on recognizing and segmenting objects from an open set of categories. ODISE~\cite{xu2023open} leverages text-to-image diffusion models to learn rich semantic representations, enabling the generation of open-vocabulary panoptic masks. FC-CLIP~\cite{yu2023convolutions} develops a segmentation model built on a shared frozen convolutional CLIP backbone, utilizing two-way open-vocabulary classification to improve performance on novel classes. OpenSeeD~\cite{zhang2023simple} unifies open-vocabulary segmentation and detection by jointly learning from segmentation and detection datasets, addressing task and data discrepancies through decoupled decoding and conditioned mask decoding.
In-context segmentation~\cite{wang2023seggpt,liu2023matcher,zhang2023personalize,li2024visual,liu2024simple,meng2024segic} enables segmentation guided by example images. SegGPT~\citep{wang2023seggpt} introduces a random coloring scheme within in-context learning to enhance model generalization. PerSAM~\citep{zhang2023personalize} employs one-shot data to generate positive-negative location priors and infuses high-level semantics from SAM~\citep{kirillov2023segment} to guide personalized segmentation. Matcher~\citep{liu2023matcher} proposes a training-free framework that utilizes vision foundation models for few-shot segmentation tasks. DINOv~\citep{li2024visual} integrates visual in-context prompting to unify referring and generic segmentation tasks, introducing a specialized encoder-decoder architecture. SINE~\citep{liu2024simple} resolves ambiguity in in-context segmentation by unifying tasks across various granularities.
Unlike these methods, COSINE unifies open-vocabulary and in-context segmentation, fostering synergistic collaboration between these two branches, and significantly enhancing the model’s open-world generalization performance.

\noindent\textbf{Vision Foundation Models.}
Leveraging contrastive and generative training, vision foundation models, when combined with large-scale single-/multi-modal datasets, exhibit remarkable generalization capabilities.
Pre-trained by masked image modeling~\cite{dosovitskiy2020image,bao2021beit,xie2022simmim,liu2024masked}, MAE~\cite{he2022masked} demonstrating strong transfer performance across downstream tasks.
DINOv2~\citep{oquab2023dinov2}, through image and patch-level discriminative self-supervised learning, learns versatile visual features applicable to a wide range of downstream tasks. 
CLIP\citep{radford2021learning} learns multi-modal representations by image-text contrastive learning and demonstrates outstanding zero-shot image classification performance.
The Segment Anything Model (SAM)\citep{kirillov2023segment} achieves impressive zero-shot, class-agnostic segmentation performance by training on a large-scale segmentation dataset.
This work aims to train an open-world segmentation model with strong generalization capabilities, leveraging the potential of existing foundation models, even under constraints in data and computational resources.

\begin{figure*}[t]
    \centering
    \includegraphics[width=0.9996\linewidth]{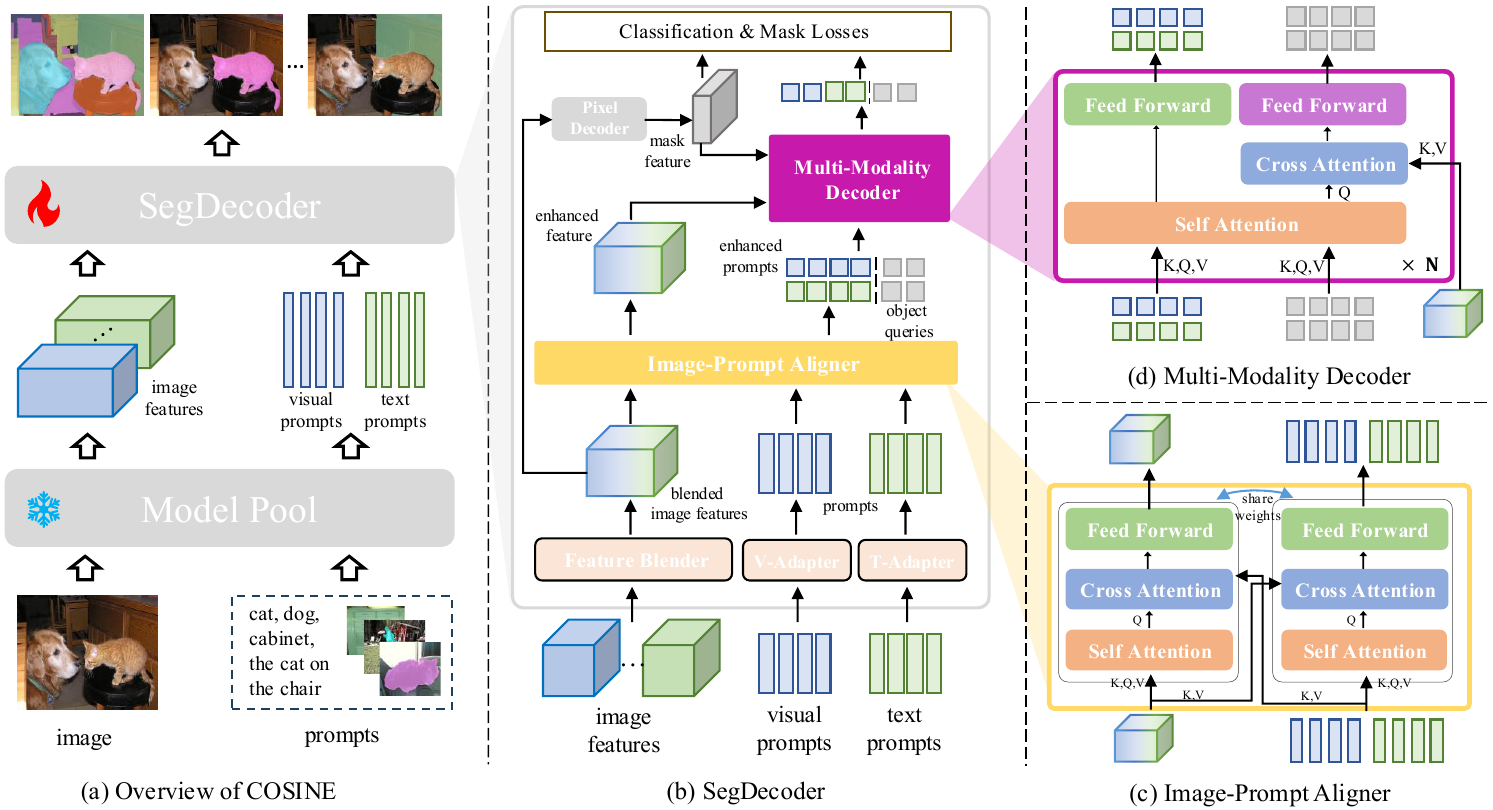}
        \vspace*{-0.2cm}
    \caption{The architecture of COSINE. (a) COSINE consists a Model Pool (\eg, DINOv2 and CLIP) used to extract image and prompt features and a SegDecoder used for unified open-world segmentation tasks. (b) SegDecoder consists a set of adapters, an Image-Prompt Aligmenter, a Pixel decoder and a Multi-Modality Decoder for modality alignment between the image and prompts, effectively enhancing open-world perception modeling. (c) and (d) show the details of Image-Prompt Aligmenter and Multi-Modality Decoder.}
    \label{framework}
\end{figure*}

\section{Method}
\label{sec:method}

We present COSINE, a unified open-world segmentation model that 
\textbf{C}onsolidates \textbf{O}pen-vocabulary \textbf{S}egmentation and \textbf{IN}-context s\textbf{E}gmentation. 
COSINE supports diverse modalities of input, such as images and text, by framing them as promptable segmentation tasks, offering powerful open-world perception capabilities.
Our goal is not to achieve state-of-the-art performance across all open-vocabulary or in-context segmentation tasks, but to validate that diverse modal information can synergistically collaborate and extend knowledge across tasks, thereby enhancing the model's modeling capabilities and generalization performance.

\subsection{Overview}

As shown in Fig.~\ref{framework}(a), COSINE follows a simple design philosophy, consisting of a Model Pool for extracting features from a target image and different modality prompts (e.g., text and example image), and a decoder-only segmentation model, named SegDecoder, which takes image and prompt features as inputs and outputs results for various segmentation tasks. 
Unlike previous segmentation methods~\cite{zhang2023simple,kirillov2023segment,li2024visual} that train all parameters of models, COSINE fixes the foundation models in the Model Pool and only trains the relatively lightweight SegDecoder. 
Our approach enables lower training overhead while unleashing the potential of the foundation models for open-world perception.

\noindent\textbf{Model Pool.} Model Pool includes different vision models (DINOv2 and CLIP vision encoder) and language models (CLIP text encoder).
The inputs of Model Pool include a target image, its image prompts and text prompts.
For the target image, we use all vision models to encode it into the image features $\mathcal{F}=\{\textbf{F}_i\}_i^P$. 
Then, we use DINOv2 to encode the images of visual prompts into image features and use the in-context masks to pool the features into prompt tokens $\mathcal{V}=\{\textbf{v}_i\}_i^{M}$. The language model is used to extract the text prompt features $\mathcal{T}=\{\textbf{t}_i\}_i^{N}$. 
The image and prompt representations extracted by different foundation models contain rich multi-modal information, and the modalities complement each other, facilitating efficient transfer to downstream perception tasks.

\subsection{SegDecoder}

The architecture of SegDecoder is shown in Fig.~\ref{framework}
(b), which include a set of adapters, an Image-Prompt Aligmenter, a Pixel decoder and a Multi-Modality Decoder. 

We deploy different adapters, including Feature Blender, V-Adapter and T-Adapter, for different features with the following objectives: (1) blending different image features, (2) aligning the feature dimensions of the image and various modality prompts, and (3) mapping the representations from different modalities to a shared space. 
Feature Blender consists of two convolutional layers that take the concatenated image features as input and outputs the blended image feature $\mathbf{F} \in \mathbb{R}^{C \times H \times W}$. Similarly, 
V-Adapter (T-Adapter) maps image (text) prompts into $\textbf{V} \in \mathbb{R}^{M \times C}$ ($\textbf{T} \in \mathbb{R}^{N \times C}$). 

\noindent\textbf{Image-Prompt Aligner.} 
The image features integrate knowledge from multiple models, while each prompt only encode knowledge from a single modality, leading to a disparity between them.
To mitigate the disparity, we employ an Image-Prompt Aligner to align and harmonize these heterogeneous representations.
As shown in Fig.~\ref{framework}(c), the Image-Prompt Aligner is implemented by a set of self-attention, cross-attention, and feed-forward networks (FFN). Through the shared network structure, the Aligner aligns the image and different modality prompts, enhancing their representations in the multi-modal space, enabling high-quality segmentation results using prompts from any modality. Specifically,
the blended image feature $\textbf{F}$ and the prompt features $\textbf{V}$ and $\textbf{T}$ are aligned in the multi-modal representation space through the Image-Prompt Aligner module:
\begin{equation}
% \begin{split}
\left<\textbf{F}^{'}, \textbf{V}^{'}, \textbf{T}^{'}\right> = Alignment\left(\textbf{F}, \textbf{V}, \textbf{T} ;\theta \right),
% \end{split}
\end{equation}
where $\textbf{F}^{'} \in \mathbb{R}^{C \times H \times W}$ is the enhanced image feature, and $\textbf{V}^{'} \in \mathbb{R}^{M \times C}$ and $\textbf{T}^{'} \in \mathbb{R}^{N \times C}$ are the enhanced prompt features. $\theta$ is the parameters of the Image-Prompt Aligner.

\noindent\textbf{Pixel Decoder.} 
The aim of the Pixel Decoder is to output high-resolution mask features. For single-scale image feature, the Pixel Decoder is implemented as a small network consisting of two transpose convolution layers. It takes the blended image feature $\textbf{F}$ as input, performs $4\times$ upsampling, and generates the mask feature $\textbf{F}_{mask} \in \mathbb{R}^{C \times H^{'} \times W^{'}}$.

\noindent\textbf{Multi-Modality Decoder.} As show in Fig.~\ref{framework}(d), we introduce a Multi-Modality Decoder to refine the different modality prompts and object queries $\textbf{Q} \in \mathbb{R}^{K \times C}$. 
Following~\cite{liu2024simple}, the Multi-Modality Decoder adopts a dual-path design, utilizing both self-attention and cross-attention~\cite{cheng2022masked} to facilitate interaction between object queries, different modality prompts, and the image features, while simultaneously refining the object queries and prompts. The process of the Multi-Modality Decoder can be summarized as follows:
\begin{equation}
% \begin{split}
\left<\textbf{Q}_{r}, \textbf{V}_{r}, \textbf{T}_{r}\right> = Decoder\left(\textbf{Q}, \textbf{V}^{'}, \textbf{T}^{'}, \textbf{F}^{'}, \textbf{F}_{mask} ;\phi \right),
% \end{split}
\end{equation}
where $\textbf{Q}_{r}$, $\textbf{V}_{r}$ and $\textbf{T}_{r}$  are refined object queries, visual and text prompts. $\phi$ is the parameters of the Multi-Modality Decoder. 
By using $\textbf{V}_{r}$ and $\textbf{T}_{r}$ as classifiers, COSINE can obtain the classification scores $\textbf{S}_v$ and $\textbf{S}_t$ for in-context segmentation and open-vocabulary segmentation. This can be illustrated in the following equation:
\begin{equation}
\begin{split}
\textbf{S}_v=\textbf{Q}_{r} \cdot {\textbf{V}_{r}}^T, \textbf{S}_v \in \mathbb{R}^{K \times M},
\\
\textbf{S}_t=\textbf{Q}_{r} \cdot {\textbf{T}_{r}}^T, \textbf{S}_t \in \mathbb{R}^{K \times N}.
\end{split}
\end{equation}
The mask results $\textbf{M}$ can be obtained via the following operation:
\begin{equation}
% \begin{split}
\textbf{M}=MLP\left(\textbf{Q}_{r}\right) \cdot {\textbf{F}_{mask}}, \textbf{M} \in \mathbb{R}^{K \times H^{'} \times W^{'}}.
% \end{split}
\end{equation}
Note that $\textbf{S}_v$ and $\textbf{S}_t$ are not computed independently. In practice, the vision and text prompts are concatenated and jointly processed to produce a unified score matrix. The separate notation of $\textbf{S}_v$ and $\textbf{S}_t$ is adopted purely for clarity of presentation.

\noindent\textbf{Multi-Scale Feature Injection.} Similar to previous approaches~\cite{li2022exploring,yu2023convolutions}, we can obtain multi-scale image features from these foundation models to enhance the performance of SegDecoder. We employ a multi-scale deformable attention Transformer~\cite{zhu2020deformable} as the Pixel Decoder and use these multi-scale image features as inputs to both the Image-Prompt Aligner and the Multi-Modality Decoder.

\subsection{Training and Inference}

\noindent\textbf{Image and Text Prompts Co-training.} 
During the training phase, we experientially maintain a 1:1 sample ratio between image and text prompts for each batch, ensuring that the model effectively balances in-context and open-vocabulary segmentation tasks within an open-world framework. For in-context segmentation, we employ two distinct sampling strategies for image prompts: (1) sampling diverse images of the same class to serve as both the target image and the image prompt, thereby enhancing the model's capability to discern target semantics, and (2) randomly cropping a single image to generate multiple views of the same object, which strengthens the model's ability to recognize objects across varying perspectives. For open-vocabulary segmentation, we sample a set of class names that encompass the target image classes as prompts. The architecture of the SegDecoder is designed to process both in-context and open-vocabulary segmentation tasks uniformly, facilitating the use of a shared loss function for optimization.
In line with prior DETR-based approaches~\cite{carion2020end,cheng2022masked,zhu2020deformable,liu2024simple}, we use bipartite matching to match predictions with ground-truth. Subsequently, we optimize the model using cross-entropy loss for classification, along with binary mask loss and Dice loss~\cite{milletari2016v} for mask prediction.

\noindent\textbf{Multi-Modal Prompts Collaborative Inference.} 
The SegDecoder architecture empowers COSINE to seamlessly process inputs from single-modality prompts, enabling it to tackle both open-vocabulary segmentation and in-context segmentation tasks. Furthermore, COSINE supports collaborative inference by leveraging prompts from multiple modalities. By integrating image and text prompts, COSINE generates mask predictions based on blended instructions. Notably, in addition to utilizing single-modality prompts, we introduce a straightforward averaging fusion mechanism to blend image prompts $\textbf{V}$ and text prompts $\textbf{T}$. This fusion strategy allows complementary information from different modalities to enhance COSINE’s generalization capabilities in open-world scenarios.

\section{Experiments}
\label{sec:exp}

\subsection{Experiments Setting}
\label{sec:setting}

\noindent\textbf{Training Data.}~We train our model with publicly available academic segmentation datasets, encompassing semantic, instance, and panoptic segmentation tasks. Specifically, we leverage datasets including:
COCO~\cite{lin2014microsoft} is a widely used dataset that supports multiple tasks, including object detection, instance segmentation, and panoptic segmentation. It comprises 80 ``things'' categories and 53 ``stuff'' categories, with a total of 118K training images and 5K validation images. Objects365~\cite{shao2019objects365} is a large-scale, high-quality dataset designed for object detection. It includes 365 object categories, 638K images, and approximately 10 million bounding boxes. 
Following the approach in~\cite{liu2024simple}, we utilize Objects365-SAM, an enhanced version of the Objects365 dataset where instance segmentation annotations are extended with SAM~\cite{kirillov2023segment}. 
In addition,  we incorporate referring segmentation datasets to endow COSINE with the capability of understanding referring expressions. Specifically, following LISA~\cite{lai2024lisa}, we utilize refCLEF, refCOCO, refCOCO+~\cite{kazemzadeh2014referitgame}, and refCOCOg~\cite{mao2016generation}, which are widely adopted benchmarks for referring expression comprehension and segmentation. More implementation details are provided in the Appendix~\ref{details}.

\begin{table*}[t]
	\centering
	\begin{center}
	\small
		\setlength{\tabcolsep}{2mm}
		\begin{tabular}{l l  c c  cc  c c c  c c  c c}
                \toprule
        \multirow{3}{*}{Methods} & \multirow{3}{*}{Venue} &\multicolumn{2}{c}{few-shot sem.} & \multicolumn{2}{c}{few-shot ins.} & \multicolumn{5}{c}{open-voc. pano.} & \multicolumn{2}{c}{open-voc. sem.} \cr
            &&\multicolumn{2}{c}{LVIS-92$^i$}&\multicolumn{2}{c}{LVIS}&\multicolumn{3}{c}{ADE20K}&\multicolumn{2}{c}{Cityscapes} & A-847 & PC-459\cr
            \cmidrule(lr){3-4}\cmidrule(lr){5-6}\cmidrule(lr){7-9}\cmidrule(lr){10-11}\cmidrule(lr){12-12}\cmidrule(lr){13-13}
            &&one-shot&few-shot&AP&APr&PQ&AP&mIoU&PQ&mIoU&mIoU&mIoU\cr
		\midrule
            \multicolumn{2}{l}{\textit{\small few-shot model}} &   &  &  &  &  &  &  & &    & & \cr
            HSNet~\cite{min2021hypercorrelation} & ICCV'21  &  17.4 & 22.9 & - & - & - & - & - & - & -  & - & -  \cr
            VAT~\cite{hong2022cost} & ECCV'22  & 18.5 & 22.7  & - & - & - & - &  - & - & - & -  & -  \cr
            DiffewS~\cite{zhu2024unleashing} & NeurIPS'24  & 31.4 & 35.4  & - & - & - & - & - & - & -  & - & -  \cr
            \midrule
            \multicolumn{2}{l}{\textit{\small in-context model}} &   &  & &  &  &  &  &  &  & &  \cr
            SegGPT~\cite{wang2023seggpt} & ICCV'23 &  18.6 & 25.4 & - & - & - & - & - & - & - & -  & -  \cr
            PerSAM-F~\cite{zhang2023personalize} & ICLR'24 & 18.4 & - & -  & - & - & - & - & - & - & -  & - \cr
            Matcher~\cite{liu2023matcher} & ICLR'24 & 33.0 & 40.0  & - & - & - & - & - &  - & - & -  & -  \cr
            SINE~\cite{liu2024simple} & NeurIPS'24 &  31.2  & 35.5  & 8.6 & 7.1 & - & - & - & - & - & -  & -  \cr
            \midrule
            \multicolumn{2}{l}{\textit{\small open-vocabulary model}} &   &  & &  &  &  &  &  & &   & \cr
            ODISE~\cite{xu2023open}  & CVPR'23 & -  & - &  - & -& 23.4  & 13.9 & 28.7 & 23.9 & -  & 11.1 & 14.5  \cr
            FC-CLIP~\cite{yu2023convolutions}  & NeurIPS'23 &  - & - & - & - & 26.8 & 16.8 & 34.1 & 44.0 & 56.2  & 14.8 & 18.2  \cr
            HIPIE~\cite{wang2023hierarchical} & NeurIPS'23 &  - & - & - & - & 22.9 & 19.0 & 29.0 & - & -   & 9.7 & 14.4  \cr
            SED~\cite{xie2024sed} & CVPR'24 &  - & -  & - & - & - & - & - & - & - & 13.9 & 22.6  \cr
            \midrule
            \multicolumn{2}{l}{\textit{\small universal model}} &   &  &  & &  &  & &  &  & \cr
            X-Decoder~\cite{zou2023generalized} & CVPR'23 &  - & - & - & - & 21.8 & 13.1 & 29.6 & 38.1 & 52.0 & 9.2 & 16.1  \cr
            UNINEXT$^{*}$~\cite{yan2023universal} & CVPR'23 &  - & - & - & - & 8.9 & 14.9 & 6.4 & - & - & 1.8 & 5.8  \cr
            OpenSeeD~\cite{zhang2023simple}  & ICCV'23  &  -  & - & - & - & 19.7 & 15.0 & 23.4 & 41.4 & 47.8 & -  & -  \cr
            DINOv~\cite{li2024visual} & CVPR'24 & -  & -  & 15.4 & 14.5 & 23.2 & 15.1 & 25.3 & - & - & -  & -  \cr
            OMG-Seg~\cite{li2024omg}   &  CVPR'24 & -  & - & - &  - & 27.9 & - & - & - & - & -  & -  \cr
            PSALM~\cite{zhang2024psalm} &  ECCV'24 &  -  & - & - & - & - & 13.9 & 24.4 & - & - & - & 14.0  \cr
            % \midrule
           COSINE$^{\dagger}$ & \multirow{2}{*}{this work} & 34.2 & 39.1 & 17.4 & 23.3 & 28.1 & 16.7 & 35.2 & 37.1 & 53.4 &  15.2 &  19.6 \cr
            COSINE &  & 35.2 & 40.7 & 20.3 & 25.8 & 31.0 & 21.1 & 35.7 & 42.0 & 56.1  & 15.6 & 19.2  \cr

            \bottomrule
		\end{tabular}
	\end{center}
        \vspace*{-0.5cm}
        \caption{Results of different open world segmentation tasks including few-shot semantic segmentation, open-vocabulary panoptic segmentation and semantic segmentation. $^{*}$ We report the performance evaluated in~\cite{wang2023hierarchical}. $^{\dagger}$ indicates the single-scale variant of COSINE. }
        % \vspace*{-0.4cm}
	\label{tab:main}
\end{table*}

\subsection{Main Results}

We simultaneously evaluate COSINE's ability to process prompts in both the visual and textual modalities. For the visual modality, we select few-shot (semantic/instance) segmentation and video object segmentation tasks, while for the textual modality, we choose open-vocabulary (semantic/panoptic) segmentation and referring segmentation tasks. We compare COSINE against task-specific expert models as well as several general-purpose segmentation models. The results, presented in Table \ref{tab:main}, demonstrate that COSINE exhibits a significant advantage in most tasks.

\noindent\textbf{Few-Shot  Semantic Segmentation.} 
Following the evaluation protocol of~\cite{liu2023matcher}, we evaluate COSINE on the LVIS \cite{gupta2019lvis} dataset for few-shot semantic segmentation. Compared with the specialized few-shot model DiffewS~\cite{zhu2024unleashing} and the in-context model, SINE ~\cite{liu2024simple}, COSINE achieves better performance in both one-shot and few-shot settings. 
COSINE decouples low-level mask features from high-level image and prompt features. Using a query-based decoder, it achieves strong contextual understanding and accurate masks. COSINE outperforms SegGPT~\cite{wang2023seggpt}, demonstrating its effective contextual reasoning. 
As shown in the first column of Table~\ref{tab:main}, COSINE even outperforms the SAM-based model, Matcher~\cite{liu2023matcher}, despite being trained on only a small amount of segmentation data.

\noindent\textbf{Few-Shot Instance Segmentation.} 
We evaluate the few-shot instance segmentation performance of COSINE on the long-tailed dataset LVIS~\cite{gupta2019lvis}, which contains over 1,000 categories. For each category, we randomly select 10 samples (or all available samples if fewer than 10 are present). We report the AP  for all categories and the APr for rare categories.
As shown in Table~\ref{tab:main}, COSINE achieves an AP of 20.3, significantly outperforming the universal model DINOv. Furthermore, COSINE achieves an APr of 25.8, demonstrating stronger generalization to rare categories compared to previous methods. These results highlight COSINE’s superior adaptability to open-world scenarios.

\noindent\textbf{Open-Vocabulary Panoptic Segmentation.} 
Following ODISE~\cite{xu2023open} and FC-CLIP~\cite{yu2023convolutions}, we evaluate COSINE on the ADE20K~\cite{zhou2019semantic} and Cityscapes~\cite{cordts2016cityscapes} datasets for open-vocabulary panoptic segmentation. We compare COSINE with specialized open-vocabulary models, including ODISE, FC-CLIP, and HIPIE~\cite{wang2023hierarchical}, as well as general-purpose models such as X-Decoder~\cite{zou2023generalized} and OpenSeeD~\cite{zhang2023simple}. As shown in Table~\ref{tab:main}, COSINE achieves outstanding performance on ADE20K, reaching a PQ of 31.0 and an AP of 21.1. Additionally, on Cityscapes, COSINE attains performance comparable to state-of-the-art methods, further demonstrating its effectiveness in open-vocabulary panoptic segmentation.

\begin{table*}[t]
    \centering
        % \vspace*{-0.3cm}
        \small
        \setlength{\tabcolsep}{4.mm}
    % {\resizebox{0.85\textwidth}{!}{
    \begin{tabular}{l l  ccc  ccc  cc}
    \toprule
    \multirow{2}{*}{{Method}} & \multirow{2}{*}{{Venue}} & \multicolumn{3}{c}{{refCOCO}} & \multicolumn{3}{c}{{refCOCO+}} & \multicolumn{2}{c}{{refCOCOg}} \cr
    \cmidrule(lr){3-5}\cmidrule(lr){6-8}\cmidrule(lr){9-10}
    && val & testA & testB & val & testA & testB & val(U) & test(U) \cr
    \midrule
    % \hline
    % \emph{traditional methods} & & & & & & & & \cr
    MAttNet~\cite{yu2018mattnet} & CVPR'18 & 56.5 & 62.4 & 51.7 & 46.7 & 52.4 & 40.1 & 47.6 & 48.6 \cr
    MCN~\cite{luo2020multi} & CVPR'20 & 62.4 & 64.2 & 59.7 & 50.6 & 55.0  & 44.7 & 49.2 & 49.4 \cr
    VLT~\cite{ding2021vision} & ICCV'21  & 67.5  & 70.5  & 65.2  & 56.3  & 61.0  & 50.1  & 55.0  & 57.7 \\
    LAVT~\cite{yang2022lavt} & CVPR'22 & 72.7 &  75.8 &  68.8 &  62.1 &  68.4 &  55.1 &  61.2 &  62.1 \cr
    CRIS~\cite{wang2022cris} & CVPR'22 & 70.5 & 73.2 & 66.1 & 62.3 & 68.1 & 53.7 & 59.9 & 60.4 \cr
    ReLA~\cite{liu2023gres} & CVPR'23 & 73.8 &  76.5 &  70.2 &  66.0 &  71.0 & 57.7 &  65.0 &  66.0 \cr
    X-Decoder~\cite{zou2023generalized} & CVPR'23 & - & - & - & - & - & - & 64.6 & - \cr
    SEEM~\cite{zou2023segment} & NeurIPS'23 & - & - & - & - & - & - & 65.7 & - \cr
    LISA~\cite{lai2024lisa} & CVPR'24 & 74.9 & 79.1 & 72.3 & 65.1 & 70.8 & 58.1 & 67.9 & 70.6 \cr
    COSINE & this work & 77.2 & 80.7 & 71.1 & 66.4 & 73.2 & 56.4 & 67.4 & 68.5 \cr

    \bottomrule
    \end{tabular}
    % }}
\vspace*{-0.3cm}
    \caption{Results of referring segmentation on refCOCO, refCOCO+ and RefCOCOg. We report the metric of cIoU.}
    \label{tab:res}
        \vspace*{-0.2cm}
\end{table*}

\begin{table}[t]
	\centering
    	\small
		\setlength{\tabcolsep}{1mm}
        % \resizebox{0.75\linewidth}{!}{
        
		\begin{tabular}{ l   l   c   c  }
        \toprule
		\multirow{2}{*}{Methods} & \multirow{2}{*}{Venue} & \multicolumn{1}{c}{DAVIS 2017} & \multicolumn{1}{c}{YT-VOS 2019}\cr
            \cmidrule(lr){3-3}\cmidrule(lr){4-4}
            & &$J\&F$&$G$\cr
		\midrule
            \multicolumn{2}{l}{\textit{\small with video data}}  &  &     \cr
            \gr{AOT   \cite{yang2021associating}} & \gr{NeurIPS'21} &  \gr{85.4} &  \gr{85.3}  \cr
            \gr{XMem   \cite{cheng2022xmem}} & \gr{ECCV'22}  &  \gr{87.7} & \gr{85.5} \cr
            \gr{DEVA \cite{cheng2023tracking}} & \gr{ICCV'23}  &  \gr{86.8} & \gr{85.5}  \cr
            \gr{Cutie  \cite{cheng2024putting}} & \gr{CVPR'24}  & \gr{88.8} & \gr{86.1}  \cr
            \midrule
            \multicolumn{2}{l}{\textit{\small without video data}}  &  &     \cr
            Painter   \cite{wang2022images} & CVPR'23 & 34.6 & 20.6  \cr
            SegGPT   \cite{wang2023seggpt} & ICCV'23 & 75.6 & 73.1  \cr
            SEEM   \cite{zou2023segment} & NeurIPS'23 & 58.9 & -\cr
            DINOv   \cite{li2024visual} & CVPR'24  & 73.3  & 52.0  \cr
            PerSAM-F   \cite{zhang2023personalize} & ICLR'24  & 76.1 & 46.6  \cr
            SINE \cite{liu2024simple} & NeurIPS'24 & 77.0 &  66.4  \cr
            % \midrule
            COSINE & \multirow{2}{*}{this work} & 76.7 & 66.0  \cr
            COSINE-FT &  & 80.2 & 70.0 \cr

            \bottomrule
		\end{tabular}
        % }
        \vspace*{-0.3cm}
        
        \caption{Results of video object segmentation on  DAVIS 2017, and YouTube-VOS 2019. \gr{Gray} indicates the model is trained on target datasets with video data.}
	\label{tab:vos}
        \vspace*{-0.2cm}
\end{table}

\noindent\textbf{Open-Vocabulary Semantic Segmentation.} 
For open-vocabulary semantic segmentation, COSINE outperforms all universal models and the majority of open-vocabulary models, including the LLM-based model PSALM~\cite{zhang2024psalm}. As shown in Table~\ref{tab:main}, COSINE achieves a mIoU of 15.6 on the A-847 dataset~\cite{zhou2017scene} and 19.2 on the PC-459 dataset~\cite{mottaghi2014role}, demonstrating its ability to recognize and segment novel categories effectively. Notably, while COSINE slightly lags behind specialized open-vocabulary semantic segmentation models such as SED~\cite{xie2024sed}, its key advantage lies in its versatility, enabling flexible support for a wide range of segmentation tasks.

\noindent\textbf{Video Object Segmentation.} 
We evaluate COSINE on video object segmentation using DAVIS 2017~\cite{perazzi2016benchmark} and YouTube-VOS 2019~\cite{xu2018youtube} datasets. As shown in Table~\ref{tab:vos}, we compare COSINE with both video-trained models (gray rows) and models that do not use explicit video data.
Video-trained models, such as Cutie~\cite{cheng2024putting} and XMem~\cite{cheng2022xmem}, achieve strong performance by leveraging temporal information and explicitly training on video datasets. COSINE, despite not using video data, achieves competitive performance without requiring video supervision, making it a more generalizable and scalable approach.
Compared with general-purpose segmentation models, COSINE achieves state-of-the-art performance among methods that do not use video data. On DAVIS 2017, COSINE obtains a $J\&F$ score of 76.7, significantly outperforming SEEM~\cite{zou2023segment} and DINOv~\cite{li2024visual}, and achieving comparable performance to PerSAM-F~\cite{zhang2023personalize} and SINE~\cite{liu2024simple}. 
We also introduce a fine-tuned variant, COSINE-FT, which improves the overall performance across both datasets by further tuning on the pure image prompt dataset.  This demonstrates that COSINE's segmentation capabilities can be significantly enhanced through fine-tuning, making it a versatile and scalable approach adaptable to both static and dynamic scene understanding.

\begin{table}[t]
	\centering
    	\small
		\setlength{\tabcolsep}{3.8mm}
    \centering
    \scalebox{0.8}
    {
        \begin{tabular}{c c c c c c c c c}
        \toprule
        \multicolumn{2}{c}{Prompt}&  \multicolumn{2}{c}{LVIS-92$^i$}  & \multicolumn{3}{c}{ADE20K} \cr
        \cmidrule(lr){1-2}\cmidrule(lr){3-4} \cmidrule(lr){5-7}
        vision & text & 1-shot & 5-shot & PQ & AP & mIoU    \cr
        \midrule
        \Checkmark & & 24.5 & 27.8 & - & - &  -  \cr
         &\Checkmark& - & - & 13.2 & 7.6 & 30.2  \cr
         \Checkmark& \Checkmark& 27.7 & 32.1 & 17.7 & 8.1 & 30.4 \cr
         \bottomrule
        \end{tabular}
    }
        \vspace*{-0.3cm}
        \caption{Effect of the interaction between visual and textual branches during Training. All models are trained for 10k steps.}
	\label{tab:modal_train}
        \vspace*{-0.2cm}
\end{table}

\begin{table}[t]
	\centering
    	\small
		\setlength{\tabcolsep}{3.8mm}
    \centering
    \scalebox{0.8}
    {
        \begin{tabular}{c c c c c c c}
        \toprule
        \multicolumn{2}{c}{Prompt}& \multicolumn{2}{c}{LVIS-92$^i$}  & \multicolumn{3}{c}{ADE20K} \cr
        \cmidrule(lr){1-2}\cmidrule(lr){3-4}\cmidrule(lr){5-7}
        vision & text & 1-shot & 5-shot & PQ & AP & mIoU    \cr
        \midrule
        \Checkmark & &35.2 & 40.7 & 23.8 & 15.8 &  26.3  \cr
         &\Checkmark& 37.8 & - & 31.0 & 21.1 & 35.7  \cr
         \Checkmark& \Checkmark& 43.1 & 45.9 & 31.4 & 21.3 & 36.3 \cr
         \bottomrule
        \end{tabular}
    }
                \vspace*{-0.3cm}
        \caption{Effect of the interaction between visual and textual branches during inference.}
	\label{tab:modal_test}
        \vspace*{-0.2cm}
\end{table}

\noindent\textbf{Referring Segmentation.} 
We test COSINE on the referring segmentation task following the evaluation protocol of LISA~\cite{lai2024lisa}, which requires further fine-tuning on the referring segmentation dataset. As shown in Table~\ref{tab:res}, COSINE not only surpasses all traditional methods and general-purpose segmentation models but also achieves performance comparable to the LLM-based model LISA~\cite{lai2024lisa}. This further validates COSINE's capability in handling long text descriptions and complex semantic understanding.

\begin{figure*}[t]
    \centering
    \includegraphics[width=1\linewidth]{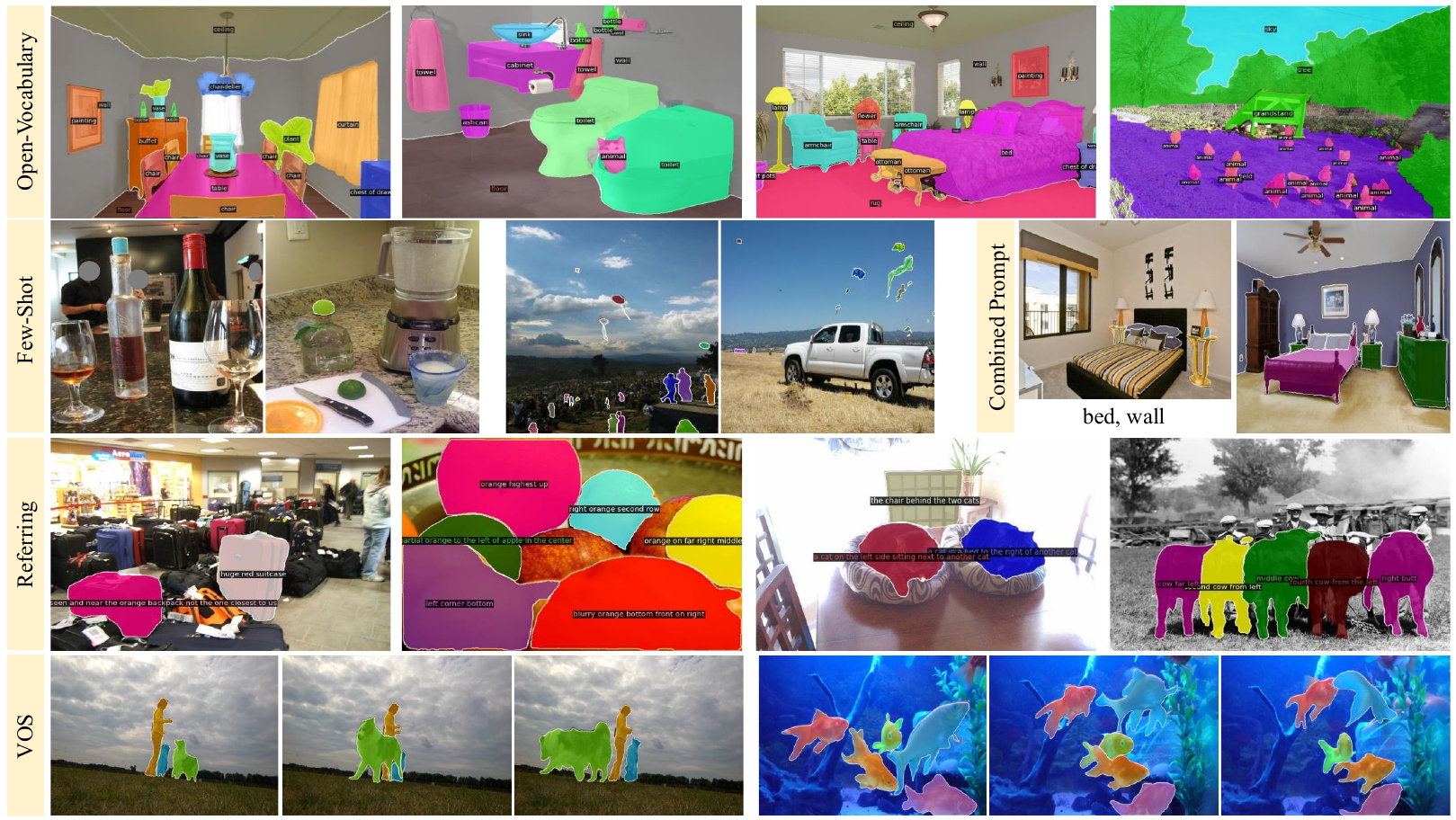}
    \vspace*{-0.6cm}
    \caption{Qualitative results. COSINE can perform various open-world segmentation tasks with different modal prompts (image and text). For few-shot segmentation, the left image is the example image and the right is the result.}
    % \vspace*{-0.3cm}
    \label{qualitative_results}
\end{figure*}

\subsection{Exploration of Visual-Textual Interaction}

As mentioned earlier, COSINE consists of a textual branch that processes text prompts and a visual branch that processes visual prompts. Few works have explored the interaction between these two modalities. COSINE is the first to investigate their interplay from two perspectives: during training and inference.

\noindent\textbf{Training Phase.} 
We aim to investigate whether jointly training both branches with multimodal data can enhance the performance of each individual branch. 
Specifically, we train the model for 10k steps using data from a single modality and multi-modality data, respectively.
As shown in Table~\ref{tab:modal_train}, we demonstrate that introducing the textual branch during training significantly improves the performance of the visual branch, and conversely, the visual branch also enhances the textual branch's performance. This finding highlights the mutual benefits of multimodal training in segmentation.

\noindent\textbf{Inference Phase.} 
We further investigate whether providing both visual and textual prompts during inference can improve model performance. We validate this hypothesis on few-shot semantic segmentation and open-vocabulary panoptic segmentation tasks, as shown in Table~\ref{tab:modal_test}. Our results reveal that for tasks originally supported by the visual branch, such as LVIS-92$^i$, incorporating an additional text prompt during inference leads to significant performance gains. Similarly, for tasks primarily supported by the textual branch, such as ADE20K, introducing a visual prompt during inference also enhances performance. 
These findings confirm that COSINE can flexibly process prompts from multiple modalities, resulting in more robust segmentation. This insight also provides valuable inspiration for future research on universal segmentation models.

\begin{table}[t]
	\centering
    	\small
		\setlength{\tabcolsep}{2.5mm}
    \centering
    \scalebox{0.7}
    {
        \begin{tabular}{l | l| c c c c c}
        \toprule
        \multirow{2}{*}{$\#$ID}&\multirow{2}{*}{Model}& \multicolumn{2}{c}{LVIS-92$^i$}  & \multicolumn{3}{c}{ADE20K} \cr
        \cmidrule(lr){3-4}\cmidrule(lr){5-7}
        && 1-shot & 5-shot & PQ & AP & mIoU    \cr
        \midrule
        0 & COSINE (Full Model) & 27.7 & 32.1 & 17.7 & 8.1 & 30.4 \cr
        \midrule
        1 & only DINOv2 Encoder & 27.7 & 32.0 & 9.7 & 4.2 & 18.6 \cr
        2 & only CLIP Encoder& 24.4 & 28.0 & 11.4 & 3.3 &  27.0\cr
        3 & w/o  Feature Blending & 24.4 & 29.0 & 9.4 & 2.6 & 26.3 \cr
        4 & w/o Image-Prompt Aligner & 26.3 & 31.3 & 12.0 & 6.4 & 30.1 \cr
        5 & w/o Prompt Refining & 27.0 & 31.2 & 16.5 & 6.6 & 30.9 \cr
        
         \bottomrule
         
        \end{tabular}
    }

        \caption{Ablation study. All models are trained for 10k steps.}
	\label{tab:ablation}

\end{table}

\begin{figure*}[t]
    \centering
    \includegraphics[width=1\linewidth]{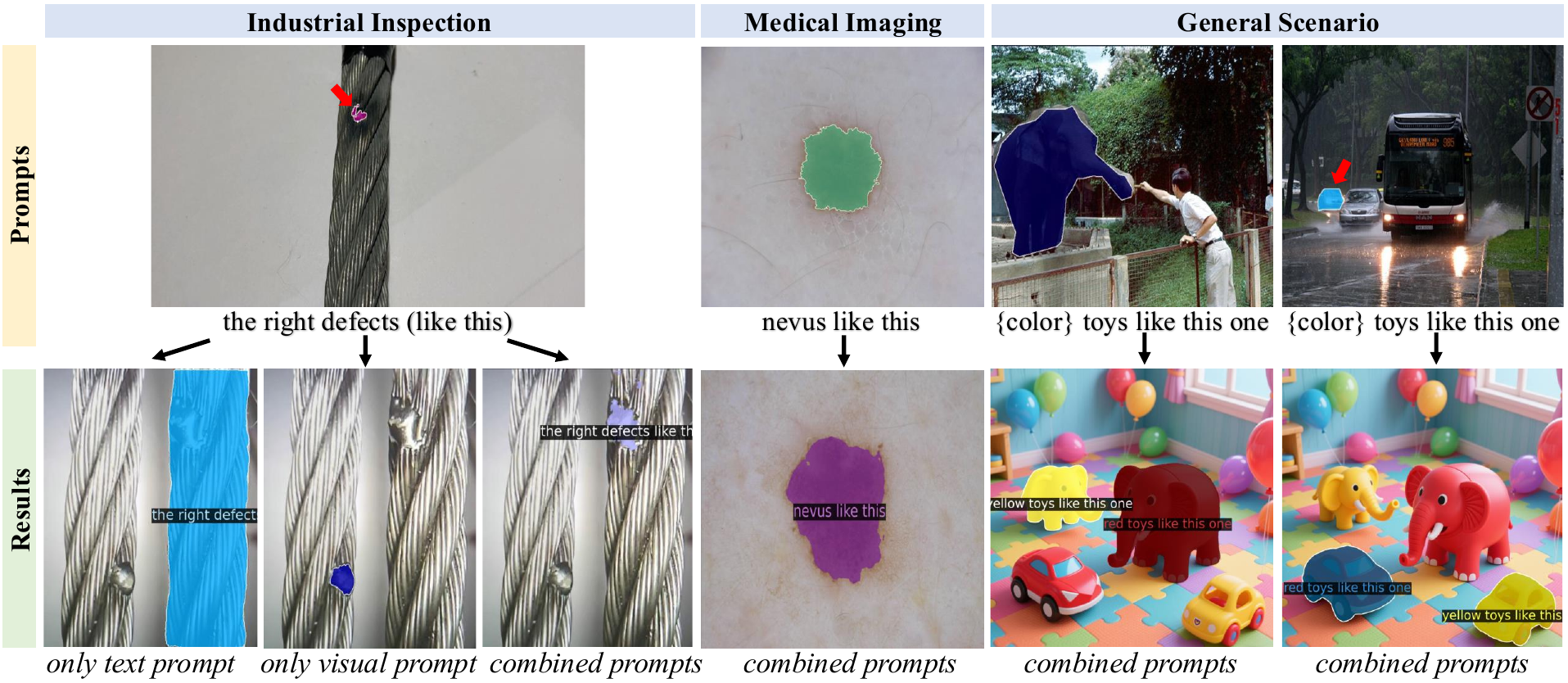}
    \vspace*{-0.6cm}
    \caption{Visualization of prompt synergy. The top row shows the input prompts, the bottom row presents the corresponding outputs.}

    \label{synergy}
\end{figure*}

\subsection{Ablation Study} 

As shown in Table~\ref{tab:ablation}, we validate the impact of each design within our model, including the Model Pool($\#1$, $\#2$), the Feature Blender($\#3$), the Image-Prompt Aligner($\#4$), and the Multi-Modality Decoder($\#5$).
Specifically, a comparison with model $\#1$ and$\#2$
reveals that different pre-trained models are informationally complementary, enhancing both in-context and open-vocabulary segmentation performance. The comparison with model $\#3$ demonstrates that merely concatenating different foundation models is insufficient. An explicit blending of representations is necessary to learn a unified multi-modal representation. Furthermore, comparisons with models $\#4$ and $\#5$ indicate that the introduced Aligner and Decoder are both simple and effective.

\subsection{Visualization Analysis}

\noindent \textbf{Qualitative Results.}
As shown in Fig.~\ref{qualitative_results}, we visualize the segmentation results under both open-vocabulary and in-context settings, as well as the results obtained by combining both types of prompts. Additionally, we provide visualizations of referring segmentation and VOS results. These results demonstrate that COSINE achieves highly accurate predictions across various modalities and granularities, highlighting its strong potential for open-world generalization.

\noindent \textbf{Prompt Synergy.}
As shown in Fig.~\ref{synergy}, COSINE can generalize to real-world domains such as industrial inspection, medical imaging. 
The industrial inspection case shows effective collaboration between visual and textual prompts, accurately segmenting targets that would produce incorrect masks under a single modality. 
These results highlight the potential of multimodal prompt synergy in COSINE for complex segmentation. By integrating visual and textual cues, COSINE captures fine-grained details that are difficult to express with a single modality.

\section{Conclusion}

In this work, we present COSINE, a unified open-world segmentation model that unifies open-vocabulary Segmentation and in-context segmentation. COSINE supports diverse modalities of input, such as images and text,  offering powerful open-world perception capabilities. 
Our exploratory analysis highlights that the synergistic collaboration between visual and textual branches enhances generalization in open-world segmentation, providing valuable insights for the research community. 
More discussion and limitations are 
provided in the Appendix~\ref{discussion}. 

\subsection*{Acknowledgments}

This work was supported by the National Key R\&D Program of China (No.\ 2022ZD0160101), the Ningbo Science and Technology Bureau 
(No.\ 2024Z291), and the National Natural Science Foundation of China (No.\ 62206244).

{
    \small
    \bibliographystyle{ieeenat_fullname}
    \bibliography{main}
}

\clearpage
\appendix

\section*{Appendix}

\renewcommand{\thefigure}{S\arabic{figure}}
\renewcommand{\thetable}{S\arabic{table}}

\section{Discussion and Limitation}
\label{discussion}

\noindent\textbf{Closed-Set Segmentation.}~To enhance open-world generalization, COSINE sacrifices some performance in closed-set scenarios. For example, on COCO, COSINE achieves a 50.6 PQ and 42.0 AP, while OpenSeeD obtains 59.5 PQ and 53.2 AP, and DINOv achieves 57.7 PQ and 50.4 AP. However, COSINE outperforms these models on unseen datasets. 
We argue that pre-trained foundation models capture a broader range of visual knowledge. Fine-tuning these models on a limited segmentation dataset can lead to performance improvements in closed-set scenarios, but it may reduce their generalization ability in unseen scenarios. Therefore, unlike existing methods, which train all model parameters, COSINE uses frozen foundation models.
We believe that the model’s ability to generalize to open-world scenarios is more critical.

\noindent\textbf{Model Pool.} The Model Pool explores a limited set of foundation models. In our preliminary experiments, we investigated the impact of the SAM encoder but did not observe significant performance improvements. Additionally, it introduced greater computational cost and constrained the input image resolution.

\noindent\textbf{Limitations.} 
Although our experiments validate that foundation models, such as DINOv2 and CLIP, exhibit complementary information, this work does not explore more advanced models with alternative training strategies, such as MLLMs~\cite{liu2023visual,zhu2023minigpt,wang2024qwen2} and diffusion models~\cite{rombach2022high,peebles2023scalable}. 
Furthermore, while COSINE leverages multiple foundation models to achieve complementary information and enhance generalization in open-world scenarios, it inevitably introduces higher computational costs. One potential solution is to distill the knowledge from different models into a single model.
These challenges will be the focus of our future work.

\noindent\textbf{Broader Impacts.}
Our approach is built upon open-source foundation models and only trains a lightweight decoder, which significantly reduces both training costs and carbon emissions. We do not anticipate any significant ethical or social concerns now.

\section{Implementation Details}
\label{details}

\noindent\textbf{Training Details.}~
We establish the frozen Model Pool by leveraging DINOv2 (ViT-L)~\cite{oquab2023dinov2} and CLIP (ConvNeXt-Large)~\cite{liu2022convnet,radford2021learning} as foundational models, while only training lightweight SegDecoder modules. Specifically, the single-scale and multi-scale variants of the SegDecoder contain 25M and 32M trainable parameters, respectively. 
The Image-Prompt Aligner has one block and the Multi-Modality Decoder has six blocks. 
All training data is converted to instance masks, and stuff classes are treated as single-instance categories. So we train only for instance segmentation, and merge instances by class at inference for semantic segmentation.
We optimize COSINE for 50K steps with a batch size of 64 using the Adam optimizer~\cite{loshchilov2017decoupled} $\left(\beta_1=0.9, \beta_2=0.999\right)$.
A linear learning rate scheduler is employed with a base learning rate of 1$e$$-$4 and a 100-step warmup phase. The weight decay is set to 0.05. For COCO and Objects365, we apply random horizontal flipping and large-scale jittering (LSJ)~\cite{ghiasi2021simple} with a random scale sampled from range 0.1 to 2.0, followed by a fixed-size crop to $896 \times 896$ for DINOv2. For CLIP, the images are resized to $1024 \times 1024$ before being inputted. For referring segmentation datasets, we only resize the images without flipping and cropping operations.

\noindent\textbf{Evaluation.} For one-shot semantic segmentation, the in-context examples are from the support sets. Like~\cite{liu2024simple}, we simply concatenate diverse image examples to accommodate the few-shot learning scenario. For few-shot instance segmentation, we randomly select 10 samples (or all available samples if fewer than 10 are present) for each category. We integrate the representations of image prompts and text prompts to form the token features.
We enhance the classification score by pooling CLIP features using the predicted masks, thereby improving the generalization capability of the model. Our method can seamlessly adopt the approach of~\cite{yu2023convolutions} to perform open-vocabulary tasks. 
For VOS, we select the first frame of the video as the image example and deploy a memory mechanism to store intermediate results, following~\cite{liu2024simple}. For referring segmentation, we adhere to the evaluation pipeline of LISA~\cite{lai2024lisa}.

\section{Additional Results}
\label{additional_results}

\begin{table}[t]
	\centering
    	\small
		\setlength{\tabcolsep}{2.2mm}
    \centering
    \scalebox{0.8}
    {
        \begin{tabular}{c c c c c c c c c c c}
        \toprule
        \multicolumn{2}{c}{Model}& \multicolumn{2}{c}{Prompt}&  \multicolumn{2}{c}{LVIS-92$^i$}  & \multicolumn{3}{c}{ADE20K} \cr
        \cmidrule(lr){1-2}\cmidrule(lr){3-4}\cmidrule(lr){5-6} \cmidrule(lr){7-9}
        DINOv2 & CLIP & vision & text & 1-shot & 5-shot & PQ & AP & mIoU    \cr
        \midrule
        \Checkmark& & \Checkmark  && 24.3 & 27.7 & - & - & -     \cr
        \Checkmark& \Checkmark & \Checkmark & & 24.5 & 27.8 & - & - &  -  \cr
         & \Checkmark& &\Checkmark& - & - & 6.6 & 2.3 & 26.8  \cr
         \Checkmark& \Checkmark& &\Checkmark& - & - & 13.2 & 7.6 & 30.2  \cr
        \Checkmark & \Checkmark& \Checkmark& \Checkmark& 27.7 & 32.1 & 17.7 & 8.1 & 30.4 \cr
         \bottomrule
        \end{tabular}
    }

        \caption{Effect of different models and training branches. All models are trained for 10k steps.}
	\label{tab:more_study}
        \vspace*{-0.2cm}
\end{table}

\noindent\textbf{Effect of different models and training branches.} 
We investigate the impact of different foundation models across various training branches. As shown in Table~\ref{tab:more_study}, DINOv2 and CLIP are commonly used foundation models for in-context and open-vocabulary segmentation tasks, respectively. The introduction of additional models further enhances performance on these tasks. When different models are jointly used for multi-modal training, complementary information is shared, enabling the models to collaborate more effectively and achieve stronger generalization performance.

\noindent\textbf{Temporal consistency assessment.} To comprehensively assess the temporal stability of video segmentation methods, we report the STB~\cite{qian2020coherent} scores on the DAVIS 2017 dataset. The results are as follows: Painter achieves an STB score of 0.91, while both SegGPT and COSINE reach 0.96, indicating significantly higher temporal consistency.

\begin{figure*}[t]
    \centering
    \includegraphics[width=0.8\linewidth]{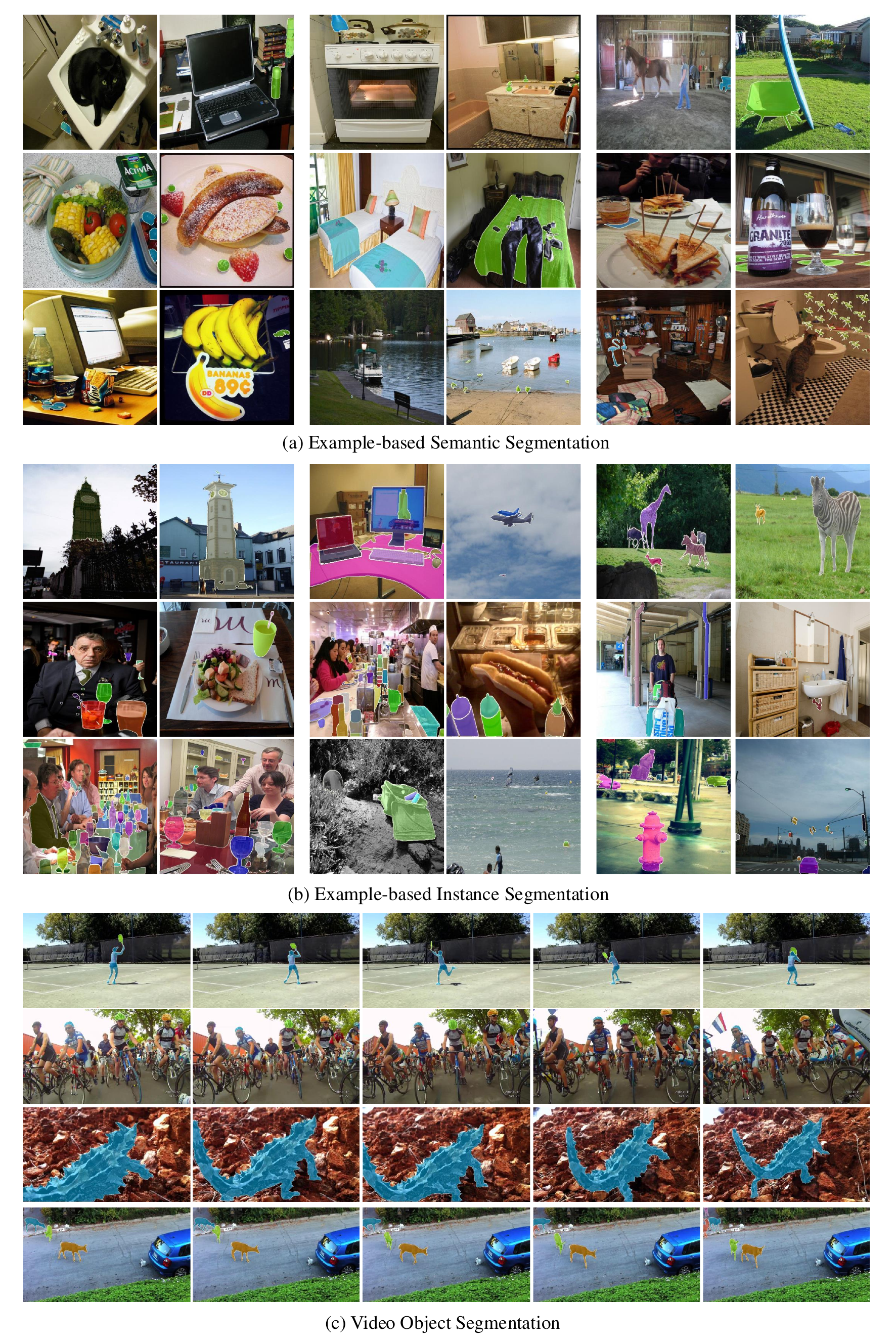}

    \caption{Visualizations of in-context segmentation tasks. (a) Example-based semantic segmentation on LVIS dataset. The left image with the blue mask is the image example, and the right image with the green mask is the result. (b) Example-based instance segmentation on LVIS dataset. We will obtain instance outputs sharing the same classes with the given image prompt. (c) Video object segmentation on the YouTuBe-VOS 2019 dataset.}

    \label{vis_ic}
\end{figure*}

\begin{figure*}[t]
    \centering
    \includegraphics[width=0.99996\linewidth]{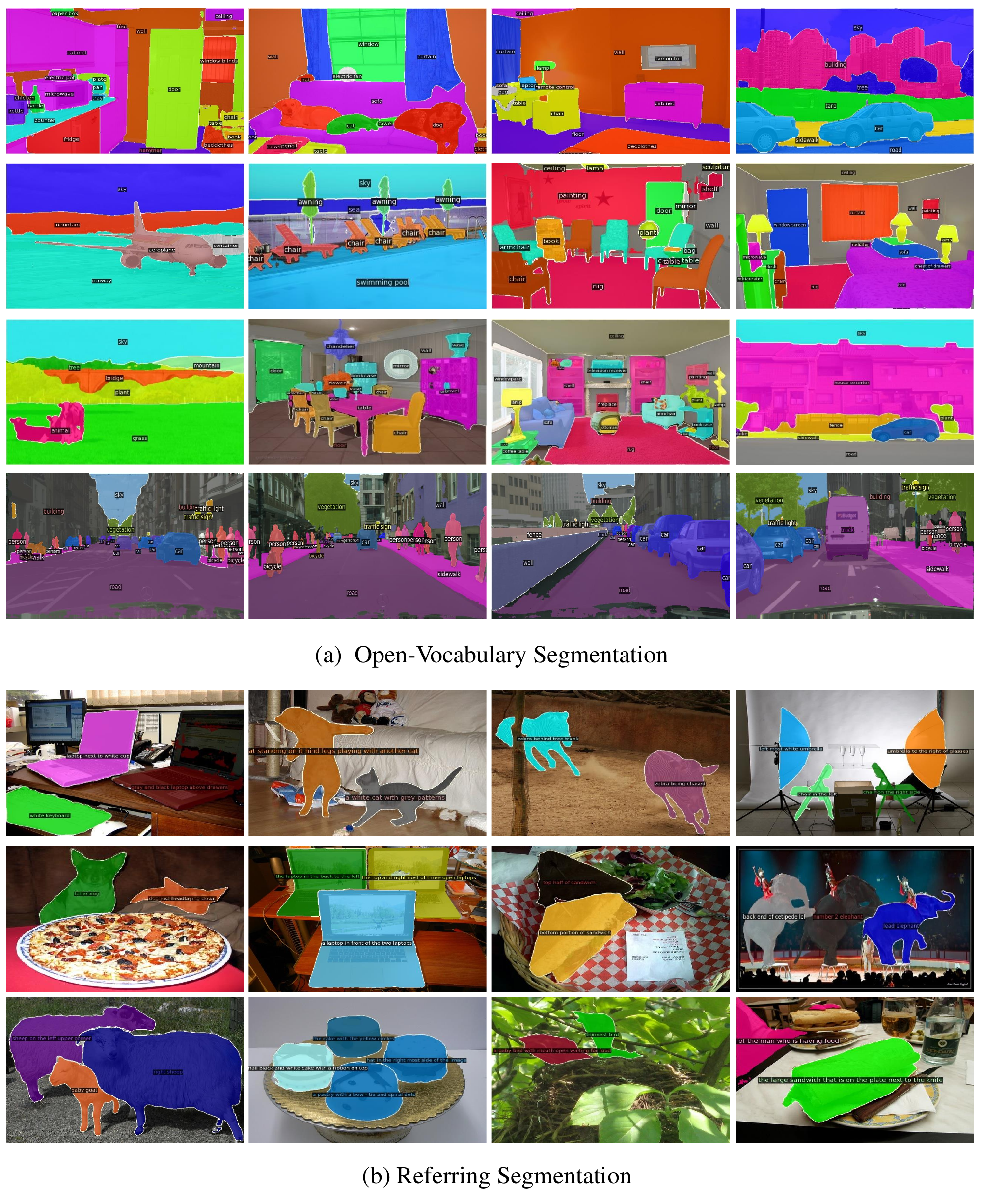}
    \caption{Visualizations of open-vocabulary segmentation and referring segmentation. }
    \label{vis_ov}
\end{figure*}

\noindent\textbf{Visualizations.} As shown in Fig.~\ref{vis_ic}, we visualize the segmentation results under in-context settings, including example-based semantic segmentation, example-based instance segmentation and video object segmentation. As shown in Fig.~\ref{vis_ov}, we visualize the open-vocabulary segmentation and referring segmentation.
These results demonstrate that COSINE achieves highly accurate predictions across various modalities and granularities, highlighting its strong potential for open-world generalization.

\end{document}